\pdfoutput=1

\documentclass[11pt,table]{article}

\usepackage[preprint]{colinglatex}

\usepackage{times}
\usepackage{latexsym}
\usepackage{graphicx}

\usepackage[T1]{fontenc}

\usepackage[utf8]{inputenc}
\usepackage{booktabs}  
\usepackage{tabularx}  
\usepackage{array}    

\newcolumntype{L}[1]{>{\raggedright\arraybackslash}p{#1}}
\usepackage{microtype}
\usepackage{multirow} 
\usepackage{url}
\usepackage{inconsolata}
\usepackage{amsmath}

\usepackage{graphicx}
\usepackage{geometry}
\usepackage{tabularx}
\usepackage{multirow}
\usepackage{lipsum}
\title{AgriCLIP: Adapting CLIP for Agriculture and Livestock via Domain-Specialized Cross-Model Alignment} 

\author{
 \textbf{Umair Nawaz},
 \textbf{Muhammad Awais},
 \textbf{Hanan Gani},
 \textbf{Muzammal Naseer},
\\
 \textbf{Fahad Khan},
 \textbf{Salman Khan},
 \textbf{Rao M. Anwer}
\\
\\
 MBZUAI, Abu Dhabi, UAE
\\
 \small{
   \textbf{Correspondence:} \href{mailto:email@domain}{umair.nawaz@mbzuai.ac.ae}
 }
}

\begin{document}
\maketitle
\begin{abstract}

Capitalizing on vast amount of image-text data, large-scale vision-language pre-training has demonstrated remarkable zero-shot capabilities and has been utilized in several applications. However, models trained on general everyday web-crawled data often exhibit sub-optimal performance for specialized domains, likely due to domain shift. Recent works have tackled this problem for some domains (e.g., healthcare) by constructing domain-specialized image-text data. However, constructing a dedicated large-scale image-text dataset for sustainable area of agriculture and livestock is still open to research. Further, this domain desires fine-grained feature learning due to the subtle nature of the downstream tasks (e.g, nutrient deficiency detection, livestock breed classification). To address this we present AgriCLIP, a vision-language foundational model dedicated to the domain of agriculture and livestock. First, we propose a large-scale dataset, named ALive, that leverages customized prompt generation strategy to overcome the scarcity of expert annotations. Our ALive dataset covers crops, livestock, and fishery, with around 600,000 image-text pairs. Second, we propose a training pipeline that integrates both contrastive and self-supervised learning to learn both global semantic and local fine-grained domain-specialized features. Experiments on diverse set of 20 downstream tasks demonstrate the effectiveness of AgriCLIP framework, achieving an absolute gain of 7.8\% in terms of average zero-shot classification accuracy, over the standard CLIP adaptation via domain-specialized ALive dataset. Our ALive dataset and code can be accessible at \href{https://github.com/umair1221/AgriCLIP/tree/main}{Github}.

\end{abstract}

\section{Introduction}

Recent years have seen success of large-scale image-text pre-training, e.g., CLIP in general zero-shot capabilities, and their widespread utility~\cite{radford2021learning}. However, the performance of these models often falters in specialized domains (such as healthcare, geo-sensing, and climate ~\cite{wang2022medclip,vivanco2024geoclip,mishra2024paperclip}) due to the presence of inherent domain gaps and the different nature of downstream tasks in specialized domains. This gap in performance has led to the curation of image-text datasets from existing domain-specific data sources for the training of expert CLIP variants~\cite{wang2022medclip,zhang2023biomedclip,xu2024addressclip}. 

However, adapting vision-text pre-training for agriculture is challenging due to two reasons. First, unlike many other fields, agriculture lacks any comprehensive image-text data sources. Existing agricultural datasets are predominantly designed for narrow tasks (e.g., disease classification) and consist only of images and task-specific information (e.g., class names), restricting their utility in vision-language pre-training. Second, most downstream agricultural tasks require fine-grained feature learning -- such as distinguishing subtle differences in rusty patches on visually similar leaves for disease classification -- where contrastive learning alone may be insufficient. 
Some previous works have utilized deep learning for agricultural tasks~\cite{bharman2022deep,farjon2023deep}, and initial efforts have been made to fine-tune large language models for the field~\cite{arshad2024ageval}. However, to the best of our knowledge, no such effort exists for vision-language pre-training for agriculture.

\begin{figure*}[hbtp]
    \centering
    \includegraphics[width=1\textwidth]{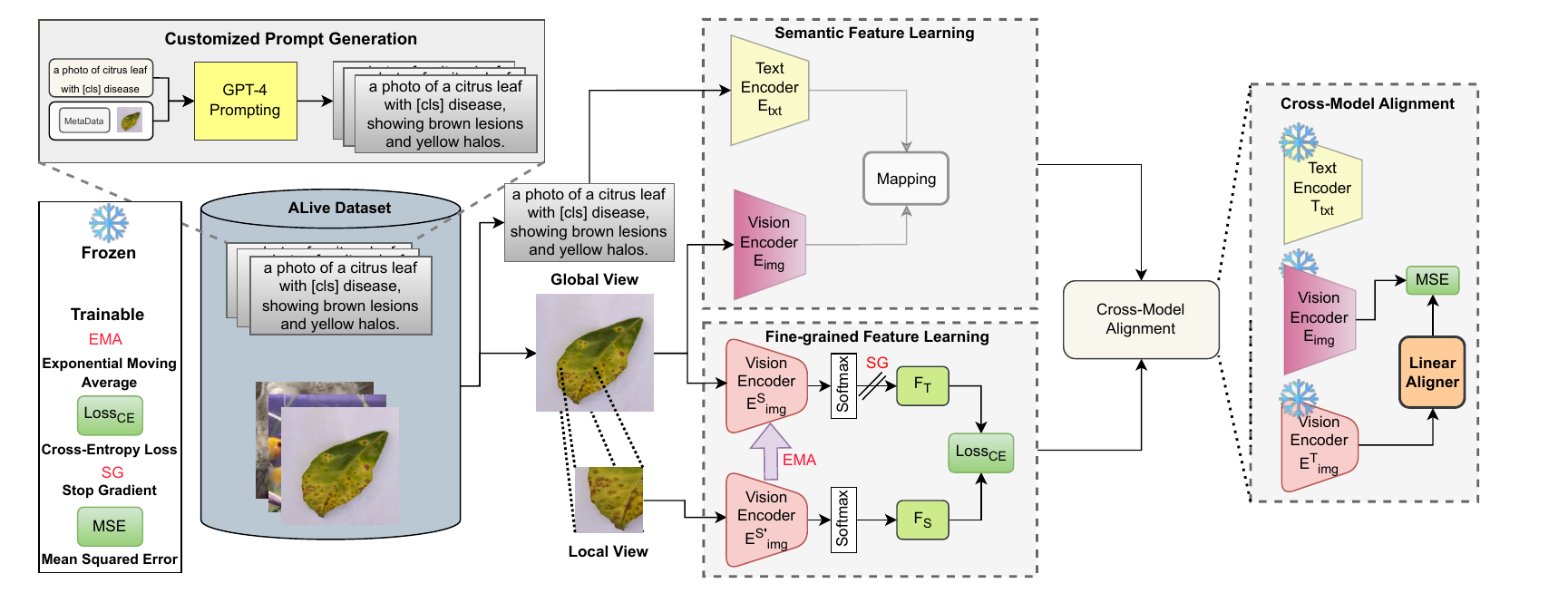}
    \vspace{-1em}
    \caption{Overview of our proposed framework, consisting of the ALive dataset and the AgriCLIP training pipeline, designed to integrate both global semantic and local fine-grained domain-specialized features. The ALive is an image-text dataset for agriculture and livestock domain that is constructed by leveraging images and their metadata to prompt GPT-4, generating customized text for each image. The AgriCLIP training pipeline consists of \textbf{Semantic Feature Learning}, where contrastive learning is utilized to train image and text encoders; \textbf{Fine-Grained Feature Learning}, using a {self-supervised} approach to train the vision encoder; and \textbf{Cross-Model Alignment}, aligning vision encoders from the previous stages to enable zero-shot generalization. 
    }
    \label{fig:AgriCLIP}
    \vspace{-1.2em}
\end{figure*}

To address the above-mentioned challenges, we introduce AgriCLIP, which comprises a large image-text dataset called ALive (Agriculture and Livestock) and a vision-language pre-training pipeline that combines the strengths of contrastive and self-supervised learning to learn both global semantic features and fine-grained visual details. This combination of a large-scale dataset and training pipeline enables the vision-language model to achieve strong downstream performance in agricultural tasks. An overview of our method is shown in Figure~\ref{fig:AgriCLIP}. Our contributions are as follows.

Our primary contribution is the creation of a large, diverse image-text dataset derived solely from vision-based agricultural datasets. To construct the ALive dataset, we carefully select 25 classification-based datasets covering crops, livestock, and fish, totaling around 600,000 images. The images span various modalities (drone, robotic, RGB), tasks (e.g., nutrient deficiency detection, plant disease identification, livestock breed classification), and environments (indoor, outdoor, and underwater). To generate corresponding text pairs for each image, we use a customized prompt generation strategy that employs metadata and class-specific information to prompt GPT4 \cite{achiam2023gpt}, producing a diverse set of descriptive text. To evaluate the model's out-of-distribution performance for diverse downstream tasks, we curate a dedicated evaluation set consisting of 300,000 images from datasets entirely disjoint from the ALive training set. 

Our second contribution is a training pipeline that combines image-text contrastive and image-only self-supervised learning to boost global semantic features with fine-grained visual details. Our training pipeline consists of three stages. In the first stage, we further pre-trained CLIP~\cite{radford2021learning} vision and language encoders on image-text pairs from the ALive dataset using contrastive learning to capture global semantic features. In the second stage, we further pre-trained a separate vision encoder with the DINO-based training~\cite{caron2021emerging} method, focusing on learning local fine-grained features crucial for downstream tasks. Consequently, we align vision encoders from the first two stages to enable zero-shot classification learning. Our experiments on 20 diverse set of downstream datasets with around 300,000 images demonstrate the efficacy of our AgriCLIP in terms of zero-shot performance.

\section{Method}
\label{sec:methodology}

As discussed earlier, popular vision-language foundational models such as CLIP and its variant have demonstrated impressive zero-shot. 
However, their applicability diminishes when applied to more specialized domains due to the inherent domain gap~\cite {udandarao2024no}.

This is likely due to being pre-trained on general-purpose images depicting everyday scenes and objects that lack the fine-grained, domain-specific examples needed for agricultural and livestock tasks, such as identifying subtle distinctions in plant diseases or detecting small variations among different crop species.

We introduce AgriCLIP (see Fig.~\ref{fig:AgriCLIP}), a framework designed to bridge the domain gap in agriculture and livestock tasks. To train AgriCLIP, we construct a large-scale image-text dataset, named ALive, for agriculture and livestock. 
AgriCLIP adapts text and vision encoders to learn discriminative, domain-specialized features followed by cross-modal alignment to obtain improved feature representation.

\begin{figure}[!t]
    \centering
    \includegraphics[width=1\linewidth]{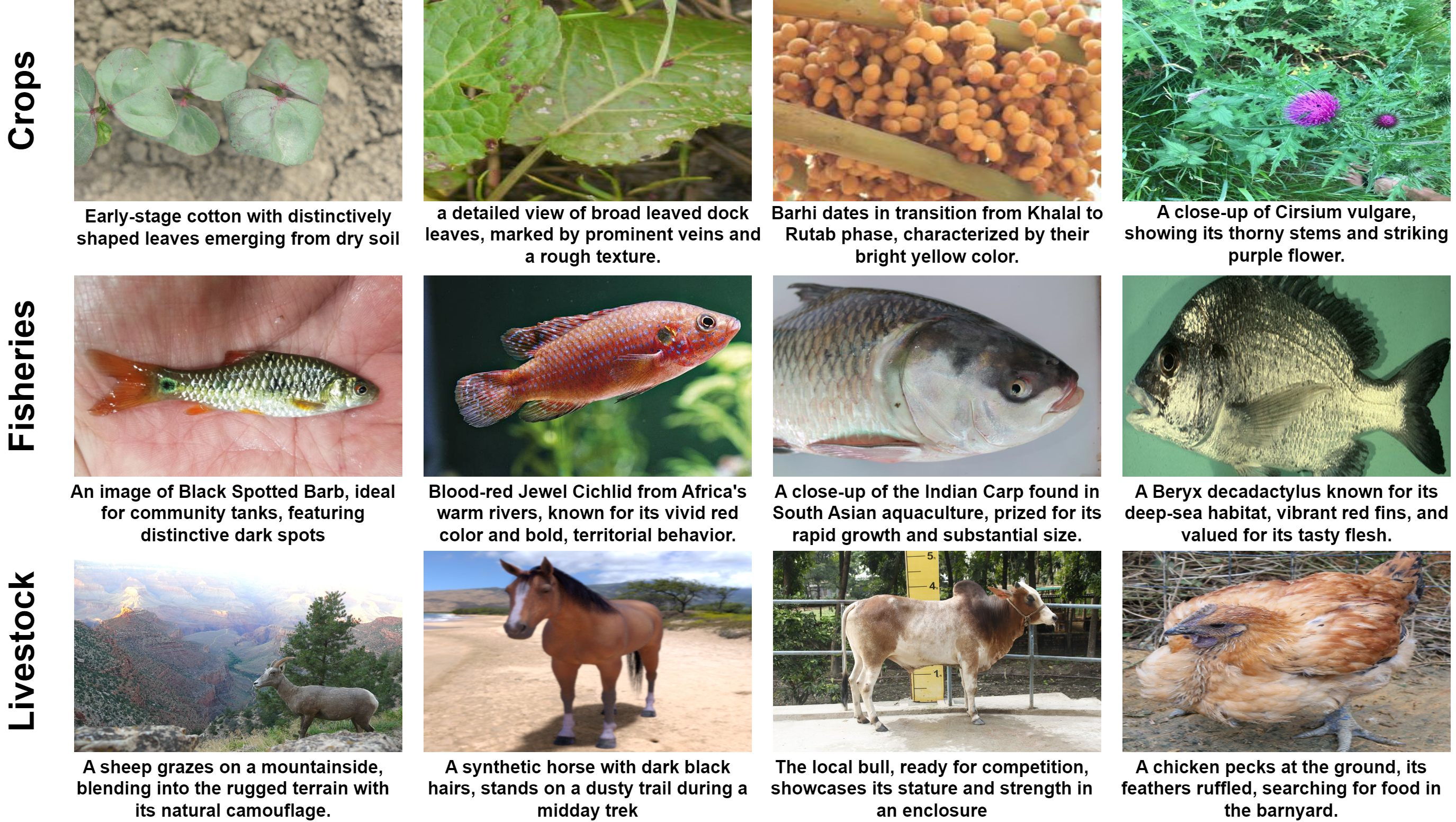}
    \caption{Example images from ALive dataset, including various crops (such as dates, crop diseases, and plant genera), diverse fish species, and samples from the livestock domain. More examples are in the appendix.}
    \label{fig:Pre-Training-Data}
    \vspace{-1.2em}
\end{figure}

\subsection{ALive Dataset for Agriculture and Livestock} 
Most existing agriculture and livestock datasets are image-only with class-level information. Here, we employ a two-step approach: first, we collect diverse agricultural and livestock data, then synthesize relevant text using dataset and class-level information, via customized prompt generation, for vision-language contrastive learning.

\begin{table*}[htbp]
\centering
\caption{Zero-shot classification performance comparison of standard CLIP, further pre-training standard CLIP on ALive dataset and the proposed AgriCLIP in a variety of downstream tasks corresponding to three domains: agriculture, livestock and fishery. 
\vspace{-0.8em}
}
\label{tab:main_table}
\resizebox{0.8\textwidth}{!}{ 
\begin{tabular}{llcccc}
\toprule
\rowcolor{gray!30}
Dataset & Downstream Tasks & \multicolumn{1}{c}{CLIP} & \multicolumn{1}{c}{CLIP Pre-Training} & \multicolumn{1}{c}{AgriCLIP} \\

\midrule
\rowcolor{gray!10}
 & \textbf{Fish} & & & \\
\midrule
Supermarket Fish \cite{ulucan2020large}   & Local Fish Classification & 1.66 & 22.57 & 41.38 \\
Aquarium Fish \cite{Aquarium28:online}    & Acquarium Fish Classification & 22.85 & 27.13 & 43.06 \\
FishDataset \cite{FishData39:online}      & Fine-grained Fish Classification & 16.04 & 32.47 & 49.23 \\
DeepFish \cite{saleh2020realistic}        & Under-Sea Fish Classification & 45.41 & 55.96 & 57.78 \\
FishNet \cite{khan2023fishnet}            & Functional Trait Prediction & 0.15 & 19.52 & 21.58 \\
Fish Freshness \cite{Fresh}               & Fish Freshness classification & 29.37 & 50.05 & 57.75 \\
Fish Species \cite{Specieclas87:online}   & Fish Species Classification & 13.68 & 25.89 & 30.21 \\
\midrule
\rowcolor{gray!10}
 & \textbf{Crops} & & & \\
\midrule
Banana Deficiency \cite{sunitha2022images} & Nutrients Deficiency Classification & 14.08 & 20.64 & 23.55 \\

Citrus Fruits \cite{sharif2018detection} & Citrus Fruit Disease Classification & 38.09 & 39.55 & 40.21 \\
Citrus Leaves \cite{sharif2018detection} & Citrus Leaves Disease Classification & 2.18 & 23.97 & 34.27 \\
Fruits Diseases \cite{kour2019plantaeK}  & Native Fruits Classification & 68.9 & 73.26 & 73.98 \\
PlantDoc \cite{PlantDoc}                  & Plant Disease Classification & 6.02 & 29.18 & 35.42 \\
Wheat Rust \cite{YELLOWRU46:online}       & Wheat Rust Classification & 34.11 & 53.45 & 67.38 \\
Bean Lesion \cite{BeanLeaf6:online}       & Bean Lesion Classification & 18.73 & 35.47 & 40.85 \\
\midrule
\rowcolor{gray!10}
 & \textbf{LiveStock} & & & \\
 \midrule
Chicken Fecus \cite{ChickenD46} & Chicken Disease Classification & 19.28 & 27.31 & 34.29 \\
CID \cite{shagor2022cid} & Local Cow Specie Classification & 05.62  & 27.52 & 49.95 \\
Cow Breed \cite{CattleBr52} & Cow Breed Classification & 31.13 & 40.23 & 44.62 \\

Animals-2 \cite{Animals176:online} & Livestock Animal Classification & 95.48 & 97.12 & 98.27 \\

Horses Breed \cite{HorseBre97} & Horses Breed Classification & 28.05 & 48.63 & 54.50 \\
\midrule
\rowcolor{gray!10} 
\textbf{Average } &  & 25.83 & 39.20 & \textbf{48.27} \\
\bottomrule
\end{tabular}
}
\vspace{-15pt}
\end{table*}


We gather 25 training datasets across crops, fish, and livestock, creating the Agriculture and Livestock (ALive) dataset with 600k images covering a wide range of conditions. This includes various crop growth stages, classifications, and different farming environments for animals and fish. 
Next, we design a customized prompt generation strategy where the text based on dataset and class-level information is leveraged to provide context and fine-grained details for each image. For instance, instead of using a generic CLIP prompt like “a photo of a boron-deficient leaf,” we craft prompts like “a photo of a leaf with boron deficiency, characterized by yellow patches and curled edges.” We then use GPT-4 \cite{achiam2023gpt} to generate diverse variation of these prompts. Table~\ref{tab:datasets-overview} in the Appendix and Figure~\ref{fig:Pre-Training-Data} present examples and details of our ALive dataset.
Next, we describe our AgriCLIP pipeline that leverages the ALive dataset.

\subsection{Learning Semantic ALive Features}
We learn global semantic domain-specialized features, via the ALive dataset, by utilizing image-text contrastive training. To this end, we adapt CLIP \cite{radford2021learning} by further pre-training it on the ALive dataset using contrastive loss \cite{he2020momentum}. The model consists of a vision encoder $E_{\text{img}}$ and a text encoder $E_{\text{txt}}$, and we align their embedding spaces by minimizing the distance between correct image-text pairs (positive pairs) and maximizing the distance for incorrect pairs (negative pairs).
The contrastive loss is defined as:

\[
L = -\log \frac{\exp(\text{sim}(u, v) / \tau)}{\sum_{k=1}^{N} \exp(\text{sim}(u, v_k) / \tau)},
\]
where $u = E_{\text{img}}(x)$ and $v = E_{\text{txt}}(y)$ are the image and text embeddings, $\text{sim}$ represents cosine similarity, $\tau$ is the temperature parameter, and N is the number of samples.
This adaptation ensures better alignment of vision and text representations for agriculture and livestock domain.

\subsection{Learning Fine-grained ALive Features}
In addition to the aforementioned domain-specialized semantic features, different agriculture and livestock problems desire capturing fine-grained visual details, crucial for tasks like identifying subtle variations in disease symptoms (e.g., small color differences in spots), for classification.

To further learn domain-specialized fine-grained features, we complement CLIP’s generalization capabilities by employing a DINO-based pre-training strategy to enhance the vision encoder ${E}^{S}_{\text{img}}$ using the ALive dataset.
The self-supervised learning-based technique \cite{caron2021emerging} excels at learning detailed visual features. Here, a student-teacher framework is employed, where two randomly augmented views of each image are processed by both models (vision transformers). The student model is trained to match the teacher’s representations, enabling it to capture both global and fine-grained details. This combined self-supervised approach enhances the model’s ability to handle domain-specialized fine-grained visual features.

\subsection{Cross-Model Alignment}  

The visual encoder ${E}^{S}_{\text{img}}$, although a powerful feature extractor, is not inherently aligned with the language encoder, and therefore lacks the zero-shot capabilities of vision-language models. 
To align domain-specialized semantic, fine-grained and text features, 
we adopt a vision-language feature alignment approach inspired by \cite{moayeri2023text}, aligning visual features ${E}^{S}_{\text{img}}$ with CLIP’s textual encoder $E_{\text{txt}}$ implicitly. Specifically, we apply a learnable affine transformation to map the output of the fine-grained visual encoder ${E}^{S}_{\text{img}}$ to the space of the semantic visual encoder $E_{\text{img}}$, effectively projecting $E^{S}_{\text{img}}$ to the same space as concept vector for text there by enabling zero-shot capabilities~\cite{moayeri2023text}. This mapping is learned by minimizing the mean squared error (MSE) between the features space of the two models.

\section{Experimental Results }

\noindent\textbf{Experimental Setup. }
For semantic visual encoder and language encoder (stage 1), we use CLIP's~\cite{radford2021learning} open-source implementation called OpenCLIP~\cite{ilharcogabriel20215143773}. For fine-grained feature vision encoder (stage 2), we utilize DINO~\cite{caron2021emerging}. It is trained with global and local crop scales of (0.4, 1) and (0.05, 0.4), respectively. The AdamW optimizer is used with a learning rate of 0.0005 and weight decay of 0.04  for a total of  100 epochs. The rest of the model settings are used as default from the DINO model.
For zero-shot evaluation, we follow the original framework utilized by CLIP~\cite{radford2021learning}. We run all our experiments on a single NVIDIA A100 GPU. 

\noindent\textbf{Downstream Tasks and Datasets.}

{ To evaluate the performance of AgriCLIP, we assemble a set of 20 datasets to test the model’s ability to generalize to unseen concepts. The evaluation set is entirely disjoint from the ALive pre-training set. 

\noindent\textbf{Results.}
We compare the performance of our AgriCLIP with both the original CLIP and its adaptation through further pre-training on the ALive dataset in Table~\ref{tab:main_table} on 20 downstream datasets. 

The original CLIP model exhibits poor performance across the 20 agriculture-related tasks, with accuracy ranging from 1.39\% to 45.41\% and an average zero-shot accuracy of 30.64\%. Further pre-training CLIP on the ALive dataset enhances performance, yielding accuracy between 20.48\% and 55.96\%, with average accuracy of 39.20\%. This gain demonstrates the impact of our ALive dataset in bridging the domain gap.
Our AgriCLIP, incorporating fine-grained feature vision encoder, further improves performance on downstream tasks. AgriCLIP achieves an overall classification score of 48.27\% over 20 datasets with an absolute gain of 9.07\% over its adapted CLIP baseline counterpart. We present more details on the experimental results and ablation studies on the impact of custom prompts, dataset size and different pre-training for fine-grained encoder in the appendix section. 

\section{Conclusion}
We present a vision-language foundational model, AgriCLIP, for agriculture and livestock domain. To facilitate model pre-training, we introduce a large-scale dataset with 600,000 image-text pairs for agriculture and livestock domain. AgriCLIP learns both semantic and fine-grained domain-specialized features for improved zero-shot classification. Experiments on 20 downstream datasets show the efficacy of AgriCLIP.
\noindent\textbf{Limitations.} The current study focuses on diverse set of classification tasks.  A potential future direction is to investigate AgriCLIP for downstream dense prediction tasks.

\bibliography{colinglatex}

\appendix

\label{sec:appendix}

\section{Related Work}

The application of AI in agriculture has been studied, with a focus on tasks such as crop monitoring \cite{wu2023challenges}, disease detection \cite{arun2023effective}, and yield prediction \cite{meena2023crop}. Traditional machine-learning approaches have relied heavily on supervised learning, requiring large amounts of labeled data \cite{kotwal2023agricultural}. However, the variability and complexity of agricultural environments often make it challenging to obtain sufficient labeled data, leading to the exploration of zero-shot learning methods.

Zero-shot learning has gained traction in recent years, with models like CLIP \cite{radford2021learning} demonstrating the ability to generalize to unseen categories by leveraging textual descriptions. CLIP's success in various domains has prompted research into its application in specialized fields such as medicine, remote sensing, and astronomy \cite{zhao2023clip,li2023rs,mishra2024paperclip}. For agriculture, efforts have been made for the specific tasks of plant disease identification in a few-shot manner, but they are mainly restricted to either a single task or limited data variability \cite{zhou2024few,zhong2020zero,sun2024few}.
Self-supervised learning models like DINO \cite{caron2021emerging} have also been explored for various applications, including visual recognition tasks \cite{li2023mask,liu2023grounding,yuen2024generalized}. DINO's ability to learn meaningful representations without labeled data makes it an engaging option for improving zero-shot learning models. 
Despite these advancements, there has been little research focused on applying these techniques to agricultural tasks. This paper aims to bridge this gap by presenting a novel approach that combines CLIP and DINO models for zero-shot classification in the agriculture domain.

\section{Extended Results}
A particularly noteworthy aspect of AgriCLIP is its ability to significantly enhance performance on datasets where CLIP exhibits notably low accuracy. For instance, on the Supermarket Fish dataset~\cite{ulucan2020large}, AgriCLIP achieves an accuracy of 40.25\%, a substantial improvement over CLIP's 1.39\%. Similarly, for the PlantDoc dataset~\cite{PlantDoc}, a large and fine-grained plant leaf disease dataset, AgriCLIP demonstrates an accuracy of 35.42\%, markedly outperforming CLIP's 6.77\%. These results highlight AgriCLIP's ability in handling diverse and challenging agricultural datasets, particularly those requiring fine-grained learning and domain-specific knowledge. 

\begin{table}[t]
\centering
\caption{Comparison between Masked Autoencoder (MAE) and DINO-based pre-training for fine-grained feature learning. DINO-based training strategy outperforms MAE significantly.}
\label{tab:mae-performance}
\resizebox{\columnwidth}{!}{%
\begin{tabular}{lcc}
\toprule
\textbf{Dataset} & \textbf{MAE} & \textbf{DINO} \\
\midrule
Supermarket Fish \cite{ulucan2020large}  &  6.20 & 41.38  \\
Aquarium Fish \cite{Aquarium28:online}  &  33.33 & 43.06\\
FishDataset \cite{FishData39:online}  & 16.21 & 49.23\\
DeepFish \cite{saleh2020realistic} & 45.41 & 57.78\\
FishNet \cite{khan2023fishnet}  & 0.11 & 21.58 \\
Banana Deficiency \cite{sunitha2022images} & 12.04 & 23.55\\
Citrus Fruits \cite{sharif2018detection} &  26.67 & 40.21\\
Citrus Leaves \cite{sharif2018detection} & 8.22  & 34.27\\
Fruits Diseases \cite{kour2019plantaeK}  & 64.55 & 73.98\\
PlantDoc \cite{PlantDoc} & 8.45  & 35.42\\

\bottomrule
Average &  24.08 & 42.04\\
\bottomrule
\end{tabular}   
}
\end{table}
\subsection{Ablations Studies}
\label{sec:discussion}

\begin{table*}[htbp]
\centering
\caption{Overview of different attributes for datasets used in ALive pre-training and downstream evaluation.}
\label{tab:datasets-overview}
\resizebox{\textwidth}{!}{%
\begin{tabular}{cp{8cm}p{4cm}p{4cm}cc}
\toprule
\textbf{Dataset} & \textbf{Tasks} & \textbf{Sensor} & \textbf{Data Variability} & \textbf{\# Images} & \textbf{\# Datasets}\\ 
\toprule
\multirow{6}{*}{Pre-Training} & Nutrients Deficiency, Weed Classification, Plant Disease Classification, Fruits Classification, Plant Specie Classification, Fish Specie Classification, Fish Abundance Classification, Cattle Classification, Cattle Breed Recognition, Horse Classification & Ground Handheld Imagery, Ground vehicle with camera attached, Ground Imagery with Fixed Camera, Ground Imagery (Using Robot), Sea Imagery & Indoor, Outdoor, Underwater, Plain Background & 603,626 & 25\\
\midrule
\multirow{4}{*}{Downstream} & Nutrients Deficiency, Weeds Classification, Plant Specie Classification, Plant Disease Classification, Cows Breed Classification, Fish Specie Classification, Functional Trait Prediction, Chicken Classification & Ground Handheld Imagery, Sea Imagery & Indoor, Outdoor, Underwater & 301,076 & 20 \\ 
\toprule
\end{tabular}%
}
\end{table*}

\noindent \textbf{Effect of Different Pre-training for Fine-grained Feature Vision Encoders. }
For fine-grained feature vision encoder, we experimented with two different self-supervised training frameworks: DINO~\cite{caron2021emerging} and Masked Auto Encoder (MAE)\cite{he2022masked}. We pre-trained both models on the ALive dataset and subsequently aligned it with CLIP, following our method. The average accuracy of 24.08\% and 44.39\% was obtained for 11 downstream tasks as shown in Table~\ref{tab:mae-performance} of Appendix. Unlike DINO, the MAE model does not perform well on downstream tasks, demonstrating suitability of DINO for handling the complexities of the ALive dataset compared with MAE.

\noindent\textbf{Effect of Custom Prompts. }
To show the effectiveness of our prompt design method, which incorporates metadata and class-specific information to synthesize text for the dataset, we perform an ablation study where we compare CLIP's fine-tuning with our prompts and generic, CLIP-style prompts. The results are shown in Table~\ref{tab:prompts}.

\begin{table}[t]
\centering
\caption{Comparison of generic prompt used by CLIP~\cite{radford2021learning} vs. customized prompts utilized in ALive dataset.}
\label{tab:prompts}
\resizebox{\columnwidth}{!}{%
\begin{tabular}{lcc}
\toprule
\textbf{Dataset} & \textbf{ Normal} & \textbf{Custom} \\
\midrule
Supermarket Fish \cite{ulucan2020large} & 22.57 & 23.45 \\
Fish Freshness \cite{Fresh} & 50.05 & 51.25 \\
PlantDoc \cite{PlantDoc} & 29.18 & 31.72 \\
Wheat Rust \cite{YELLOWRU46:online} & 53.45 & 54.82 \\
Bean Lesion \cite{BeanLeaf6:online} & 38.47 & 39.58 \\
Chicken Fecus \cite{ChickenD46} & 25.24 & 27.31 \\
CID \cite{shagor2022cid} & 17.52 & 19.28 \\

Average &  45.23  &  46.55 \\


\bottomrule
\end{tabular}
}
\vspace{-1.2em}
\end{table}

\noindent\textbf{Impact of Increasing Size of the Dataset. }
To understand effect of adding more data, we perform an ablation by increasing dataset size. To this end, the ALive dataset is expanded (ALive++) to nearly 900,000 images, encompassing a broader range of agricultural and livestock scenarios by incorporating segmentation, tracking, and detection datasets. This enhanced dataset is used solely to pre-train fine-grained feature vision encoder (stage 2) in a self-supervised manner using DINO~\cite{caron2021emerging}. After pre-training, its features are then aligned with CLIP, following stage 3 of our method. The results are shown in Table~\ref{tab:Alive-impact}, demonstrating the benefits of a larger and more varied dataset.

\begin{table}[t]
\centering
\caption{Impact of increasing size of ALive dataset.}
\label{tab:Alive-impact}
\resizebox{\columnwidth}{!}{%
\begin{tabular}{lcc}
\toprule
\textbf{Dataset} & \textbf{ALive} & \textbf{ALive++} \\
\midrule
Supermarket Fish \cite{ulucan2020large}  & 38.82 & \textbf{38.87} \\
Aquarium Fish \cite{Aquarium28:online}   & 39.14 & \textbf{42.28} \\
FishDataset \cite{FishData39:online}     & 44.76 & \textbf{52.77} \\
DeepFish \cite{saleh2020realistic}       & 57.78 & 53.43 \\
FishNet \cite{khan2023fishnet}           & 20.87 & \textbf{23.57} \\
Banana Deficiency \cite{sunitha2022images} & 20.64 & \textbf{25.34} \\
Citrus Fruits \cite{sharif2018detection} & 36.79 & \textbf{41.88} \\
Citrus Leaves \cite{sharif2018detection} & 37.87 & 37.74 \\
Fruits Diseases \cite{kour2019plantaeK}  & 70.19 & \textbf{78.58} \\
PlantDoc \cite{PlantDoc}                 & 33.48 & \textbf{38.62} \\

\bottomrule
Average & 41.87 & 43.90 \\

\bottomrule
\end{tabular}
}
\vspace{-1em}
\end{table}




\section{Extended Dataset Details}
Extended detail for the pre-training and downstream datasets are shown in Table~\ref{tab:datasets-overview}, including tasks, different sensors used for the collection of data, data variability, number of datasets, and total number of images. In Table~\ref{tbl:customexamples}, we demonstrate more examples of the customized prompts, we constructed for ALive. Also, Figure~\ref{fig:abl_fig} shows more examples of the ALive dataset.

\begin{table*}[hbtp]
\centering
\caption{Customized prompts for image descriptions used in ALive dataset. }
\label{tbl:customexamples}
\resizebox{1\textwidth}{!}{
\begin{tabular}{@{}cp{0.7\linewidth}@{}}
\toprule
\textbf{Prompt}                                & \textbf{Customized Prompts using GPT4} \\ \midrule
\multirow{4}{*}{\parbox{0.25\textwidth}{\centering a photo with prickly acacia weed specie}} & Prickly Acacia invasion in agricultural regions, an urgent weed control case \\
                                                & Impact assessment of Prickly Acacia on crop health and soil quality \\
                                                & Mapping Prickly Acacia spread in critical farming areas \\
                                                & Prickly Acacia detection in crop fields, potential for significant yield loss \\ \midrule
\multirow{4}{*}{\parbox{0.25\textwidth}{\centering a photo of rice plant leaf with nitrogen deficiency}} & A leaf displaying the yellowing characteristics of nitrogen deficiency \\
                                                & Early signs of nitrogen deficiency captured in a leaf \\
                                                & A close-up of a leaf suffering from lack of nitrogen \\
                                                & A photo of rice leaf having yellowish patterns due to nitrogen deficiency \\ \midrule
\multirow{4}{*}{\parbox{0.25\textwidth}{\centering a photo of early stage black nightsade leaves in the field}} & Black Nightshade presence in crop fields, known for its competitive nature \\
                                                & Monitoring aggressive Black Nightshade weed among vegetable crops \\
                                                & Impact of Black Nightshade on crop fields, with a focus on containment strategies \\
                                                & Field analysis of Black Nightshade weed’s effect on adjacent crops \\ \midrule
\multirow{4}{*}{\parbox{0.25\textwidth}{\centering a photo of fish from Freshwater Eel specie}} & Freshwater Eel, a species known for its elongated, snake-like body \\
                                                & Capturing the elusive Freshwater Eel during its nocturnal activity \\
                                                & The mysterious life of Freshwater Eels, seen here in a creek \\
                                                & Freshwater Eel in a clear stream, showcasing its sleek body \\ \midrule
\multirow{4}{*}{\parbox{0.25\textwidth}{\centering a photo of fish from Big Head Carp specie}} & Big Head Carp, a large species with a distinctive large head \\
                                                & Observing the feeding habits of Big Head Carp in a river setting \\
                                                & Big Head Carp, often found in river basins, impacting local ecosystems \\
                                                & A snapshot of Big Head Carp, focused on its unique head structure \\ \midrule
\multirow{4}{*}{\parbox{0.25\textwidth}{\centering a photo of a cow from Jersey breed}} & A Jersey cattle, renowned for its rich, creamy milk \\
                                                & The small yet robust Jersey cow, ideal for boutique dairy products \\
                                                & Jersey cattle in a dairy setting, noted for high butterfat content in its milk \\
                                                & A serene Jersey cow, a favorite among small-scale dairy farmers \\ \midrule
\multirow{4}{*}{\parbox{0.25\textwidth}{\centering a photo of a sheep}} & A sheep grazing peacefully, a staple of pastoral agriculture \\
                                                & Detailed capture of a sheep’s wool, essential for textile production \\
                                                & A flock of sheep on a sunny day, a vital resource for farmers \\
                                                & Sheep in a meadow, representing sustainable agricultural practices \\ \bottomrule
\end{tabular}
}
\end{table*}

\begin{figure*}
    \centering
    \includegraphics[width = \textwidth]{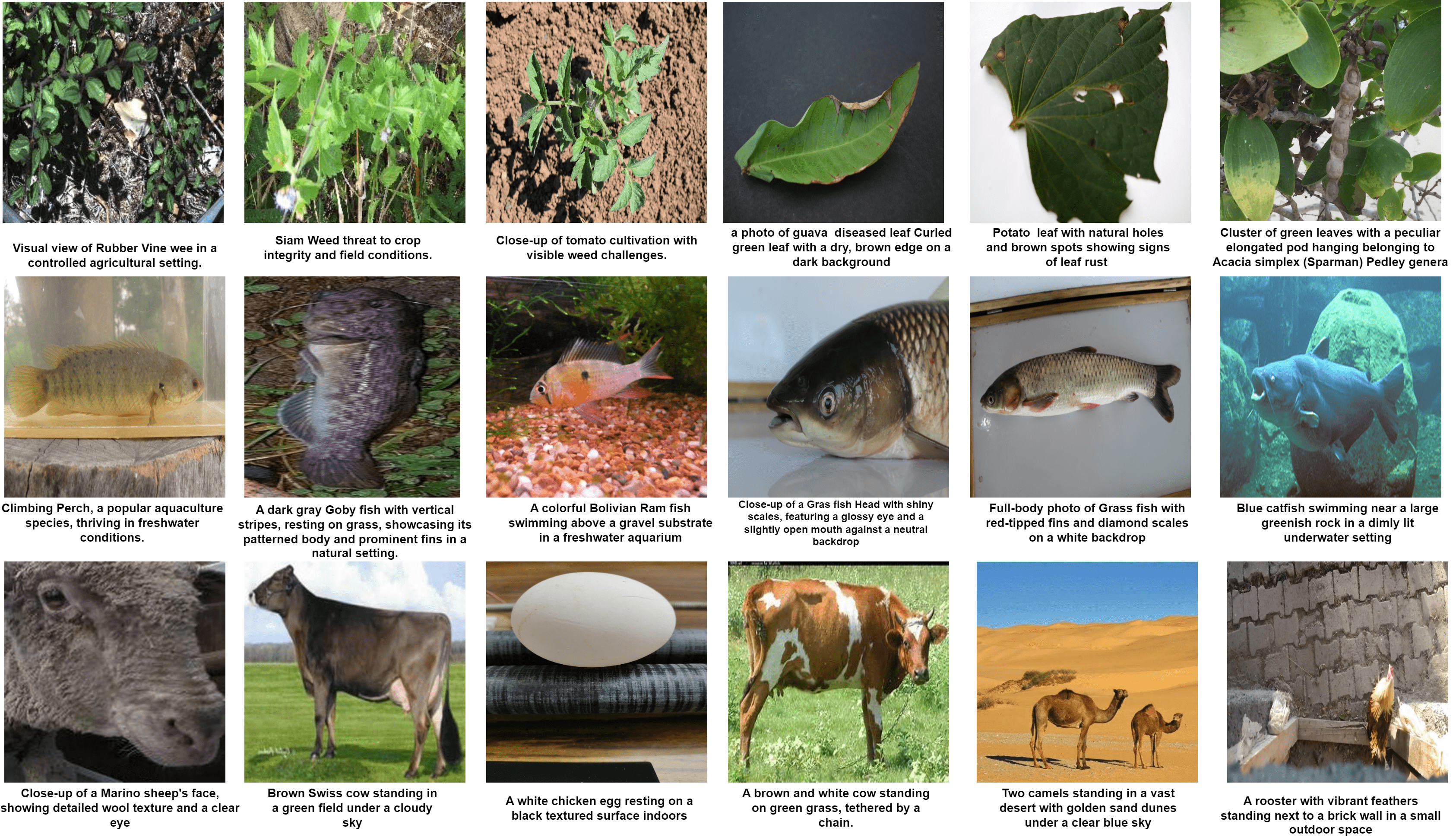}
    \caption{Some more examples of the ALive dataset}
    \label{fig:abl_fig}
\end{figure*}

\end{document}